%% file: main.tex
\documentclass{article}
\usepackage[dblblindworkshop, final]{neurips_2026}
\workshoptitle{TS-LIMITS}

\usepackage[utf8]{inputenc}
\usepackage[T1]{fontenc}
\usepackage{hyperref}
\usepackage{url}
\usepackage{graphicx}
\usepackage{wrapfig}
\usepackage{caption}
\usepackage{subcaption}
\usepackage{booktabs}
\usepackage{amsmath}
\usepackage{amssymb}
\usepackage{xspace}
\usepackage{microtype}
\usepackage{xcolor}
\usepackage{tikz}
\usetikzlibrary{arrows.meta}

\input{defines.tex}

\title{In-Context Time Series Classification\\
with Random Convolutional Features}
\author{%
  Joscha C\"uppers \\
  CISPA Helmholtz Center\\ for Information Security \\
  \texttt{joscha.cueppers@cispa.de} \\
  \And
  Jilles Vreeken \\
  CISPA Helmholtz Center\\ for Information Security \\
  \texttt{jv@cispa.de}
}

\begin{document}

\maketitle

\input{abstract} \input{introduction} \input{related} \input{method} \input{evaluation} \input{resource_experiments} \input{conclusion}

\bibliographystyle{plainnat} \bibliography{references}

\input{appendix}

\end{document}

%% file: defines.tex
\newcommand{\method}{\textsc{MASHT}\xspace}  \newcommand{\mr}{{MultiRocket}\xspace} \newcommand{\hydra}{{Hydra}\xspace} \newcommand{\tabpfnIII}{TabPFN-3\xspace} \newcommand{\hcII}{HIVE-COTE 2.0\xspace}%

%% file: abstract.tex
\begin{abstract}
	Time series classification is central to domains such as medical signal analysis, industrial monitoring, and sensor-based activity recognition, where class information manifests as localized shapes, specific frequencies, temporal shifts, or complex cross-channel interactions. Random convolutional transforms capture these diverse patterns by converting time series into rich, fixed-dimensional feature representations that can be processed by standard tabular classifiers. While these representations are traditionally paired with simple linear models, we investigate whether a pretrained tabular foundation model can exploit them more effectively and how its performance depends on the available data and inference budget.

	We propose \method, a pipeline that combines \mr and \hydra features with an in-context tabular foundation model. Our approach uses a pretrained tabular foundation model to bypass task-specific model training, requiring only feature extraction and direct inference. Extensive experiments demonstrate that \method matches state-of-the-art time series classification baselines on univariate tasks, achieving a lower average rank than \hcII. On multivariate datasets, \method remains highly competitive with the strongest reference methods. Controlled resource experiments show that compact feature tables retain most of the accuracy at substantially lower runtime, while TabPFN outperforms a matched linear baseline across the evaluated label budgets on univariate tasks. These results highlight practical trade-offs between predictive performance, labeled data, and inference cost. Code is available at: \url{https://github.com/joschac/masht}

\end{abstract}

%% file: introduction.tex
\begin{figure*}[h!]
	\centering
	\includegraphics[width=\textwidth]{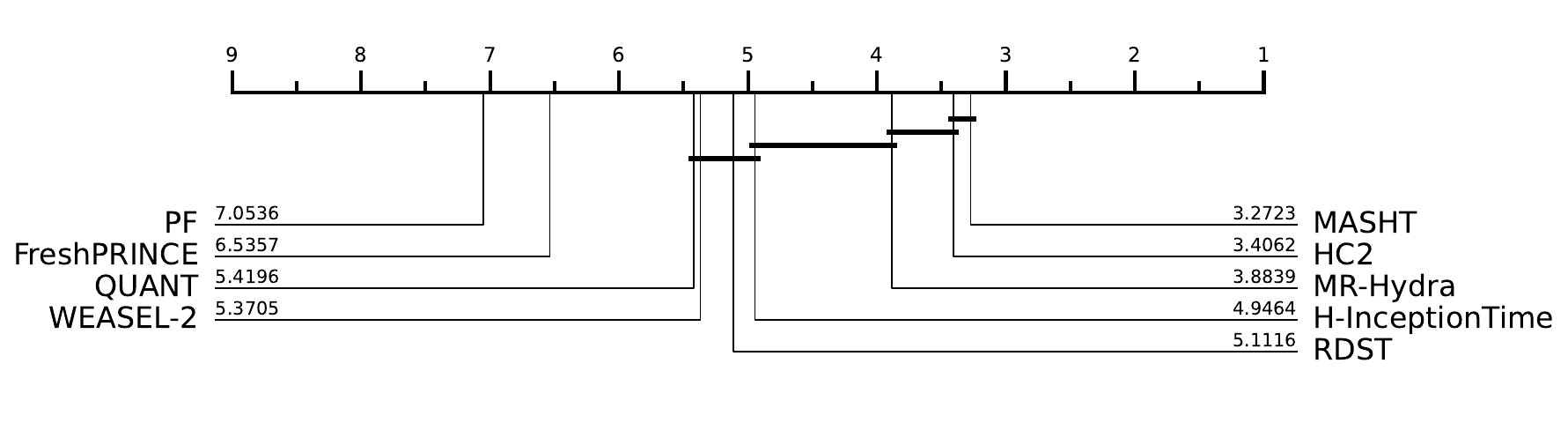}
	\caption{Critical-difference diagram for accuracy on UTF-112. Methods connected
		by a horizontal bar are not significantly different under the Wilcoxon--Holm
		comparison.}
	\label{fig:utf112-cd-accuracy}
\end{figure*}

\section{Introduction}

Time series classification (TSC) requires identifying temporal patterns that distinguish classes. Its central difficulty is representational: class information may be expressed as local shapes, frequencies, phase-independent motifs, level changes, or interactions between channels \citep{bagnall2017bakeoff,fawaz2019deep}. Highly accurate ensemble methods such as HIVE-COTE 2.0 cover many such representations, but they often achieve this breadth at substantial computational cost \citep{middlehurst2021hivecote2}. In settings with limited labeled data and tight inference budgets, the question is how much of this accuracy can be retained with fewer resources.

Random convolutional transforms offer a complementary design point. ROCKET \citep{dempster2020rocket} showed that fixed convolutional kernels paired with a simple classifier can classify accurately and efficiently. MiniRocket and MultiRocket made this family faster and more accurate \citep{dempster2021minirocket,tan2022multirocket}, while Hydra added competing convolutional count features with a dictionary-like interpretation \citep{dempster2023hydra}.

The downstream classifiers for these transformed tables are typically simple linear models, such as Ridge. They offer speed and robustness, but they fail to capture nonlinear interactions between the transformed features. In contrast, tabular foundation models \citep{hollmann2025tabpfn} are pretrained to perform in-context prediction on tabular data without any downstream training. This suggests a modular route to TSC: use specialized temporal transforms to build tabular features, then use a pretrained tabular foundation model for dataset-conditioned classification.

In this paper, we introduce \method{} (\textbf{M}ultiRocket \textbf{A}nd \textbf{S}tacked \textbf{H}ydra \textbf{T}ransformed), a modular pipeline that connects efficient temporal representations with pretrained, in-context tabular classification to improve time series classification (TSC). By concatenating random convolutional representations from MultiRocket and Hydra, the pipeline feeds a comprehensive feature table into TabPFN-3 \citep{grinsztajn2026tabpfn3} for classification. We evaluate \method{} against established univariate and multivariate TSC baselines across 112 univariate and 71 multivariate datasets using aggregate metrics, mean ranks, and Holm-corrected pairwise Wilcoxon significance tests. We show that \method{} significantly outperforms MR-Hydra on the univariate benchmark and remains competitive with the leading methods on the multivariate benchmark. In controlled resource experiments, we compare TabPFN with Ridge using identical training cases and features and find that TabPFN achieves higher accuracy on both benchmarks and across the evaluated label budgets on UTF-112. We further vary the feature budget and the number of TabPFN ensemble members and show that, with eight ensemble members, compact feature tables retain nearly all the accuracy of wider representations at substantially lower end-to-end runtime.

%% file: related.tex
\section{Related Work}

Our work connects two research directions: efficient representations for time series classification and in-context prediction with tabular foundation models. We first situate the approach within the broader TSC literature and its benchmarking practices, then review the random convolutional and dictionary-like transforms used to construct fixed tabular features, and finally discuss tabular foundation models as downstream classifiers for these representations.

\subsection{Time Series Classification}

Time series classification (TSC) methods are commonly grouped into distance-, dictionary-, interval-, shapelet-, kernel-, feature-, and deep-learning-based approaches \citep{bagnall2017bakeoff,fawaz2019deep}. HIVE-COTE 2.0 combines multiple representations in a large hierarchical ensemble and remains a prominent high-accuracy reference point \citep{middlehurst2021hivecote2}. The broader benchmarking literature provides the context for both univariate and multivariate comparisons \citep{dau2019ucr,ruiz2021bakeoff,middlehurst2026multiverse}.

\subsection{Efficient Time Series Feature Transforms}

ROCKET transforms each series with random convolutional kernels and summarizes the responses using pooling operators \citep{dempster2020rocket}. MiniRocket restricts the kernel construction to obtain an almost deterministic and much faster transform \citep{dempster2021minirocket}. MultiRocket expands this design with multiple pooling operators and features from both the raw series and its first-order difference \citep{tan2022multirocket}. Hydra organizes convolutional kernels into competing groups and records which kernels produce the strongest responses \citep{dempster2023hydra}. Its features therefore capture dictionary-like counts while retaining efficient convolutional computation. Prior work reports that concatenating Hydra with ROCKET-family features improves accuracy, motivating our choice of a combined representation. QUANT and WEASEL 2.0 provide complementary interval and dictionary perspectives on scalable TSC \citep{dempster2024quant,schaefer2023weasel}.

\subsection{Tabular Foundation Models}
Tabular foundation models treat supervised prediction as inference conditioned on a labeled training table. Rather than fitting a new model from scratch for each dataset, they use a transformer pretrained on many synthetic tasks to learn a general prediction procedure \citep{hollmann2025tabpfn,qu2026tabiclv2}. At inference time, labeled examples are provided as context for predicting unseen samples, without updating the model's pretrained weights. TabPFN-3 extends this line to greater scale and faster inference \citep{grinsztajn2026tabpfn3}.

Our work sits between these literatures. Prior TSC work studies strong temporal representations, and prior tabular foundation-model work studies generic tables. We test whether a pretrained tabular classifier can improve classification with fixed random-convolutional TSC representations after MultiRocket and Hydra encode temporal structure.

Concurrent work has also explored closely related uses of tabular foundation models for time-series classification. \textsc{RocketPFN} combines ROCKET random-convolutional features with TabPFN for training-free time-series classification, following a similar approach of first encoding temporal structure into a fixed feature table and then performing in-context prediction with a pretrained tabular model \citep{orourke2026rocketpfn}. \textsc{TS2TabPFN} likewise studies explicit time-series feature extraction followed by TabPFN \citep{merlin2026ts2tabpfn}. In contrast, \textsc{TSPFN} incorporates temporal structure directly into the foundation model through dedicated temporal representations and positional information, rather than relying solely on an external feature transform \citep{stympopper2026tspfn}.

%% file: method.tex
\section{Method}

\method{} follows a two-stage design. It first maps each input series to a fixed-dimensional feature vector by concatenating complementary MultiRocket \citep{tan2022multirocket} and Hydra \citep{dempster2023hydra} representations. It then supplies the transformed training examples and query series to the pretrained TabPFN-3 \citep{grinsztajn2026tabpfn3} classifier for in-context prediction, without learning a task-specific temporal model end to end. This separation lets the random convolutional transforms encode temporal structure while the tabular foundation model captures relationships among the resulting features. We next formalize the prediction problem before describing the pipeline and its feature-budgeted implementation.

\subsection{Problem Setting}

Let \(\mathcal{D}_{\mathrm{train}}=\{(\mathbf{x}_i,y_i)\}_{i=1}^{n}\) be a labeled TSC dataset. A univariate series is \(\mathbf{x}_i\in\mathbb{R}^{L}\) and a multivariate series is \(\mathbf{x}_i\in\mathbb{R}^{C\times L}\), where \(C\) is the number of channels and \(L\) is the series length. The goal is to estimate class probabilities \(p(y\mid \mathbf{x}_{*},\mathcal{D}_{\mathrm{train}})\) for an unseen series \(\mathbf{x}_{*}\).

\subsection{Pipeline}

Our pipeline computes a MultiRocket representation \(\phi_{\mathrm{MR}}(\mathbf{x})\), a Hydra representation \(\phi_{\mathrm{H}}(\mathbf{x})\), and then classifies their concatenation:
\begin{equation}
	\begin{split}
		z(\mathbf{x})                 & = [\phi_{\mathrm{MR}}(\mathbf{x});\phi_{\mathrm{H}}(\mathbf{x})], \\
		\hat{p}(y\mid \mathbf{x}_{*}) & =
		f_{\mathrm{TabPFN\text{-}3}}(z(\mathbf{x}_{*});\mathcal{D}_{z}),
	\end{split}
\end{equation}
where \(\mathcal{D}_{z}=\{(z(\mathbf{x}_i),y_i)\}_{i=1}^{n}\) is the transformed training set.

\subsection{Feature Extraction}

In this section, we describe how MultiRocket \citep{tan2022multirocket} and Hydra \citep{dempster2023hydra} extract features from time series data. For both transforms, we specify a target feature budget rather than a strict upper bound, since features are generated in discrete blocks.

MultiRocket applies fixed convolutional kernels over both the raw input and its first-order difference. Multiple pooling operators summarize the kernel responses through proportions, positive magnitudes, positive-response locations, and stretch lengths \citep{tan2022multirocket}. We split the requested MultiRocket budget across raw and differenced inputs, with four pooling features per kernel.

Hydra \citep{dempster2023hydra} uses convolutional kernels at different dilations; let $d$ denote the number of dilation values. For each dilation, $g$ groups are used, with each group consisting of $k$ random convolutional kernels. Each kernel produces two features. Within each group, the $k$ kernels compete at every time point: Hydra identifies the kernel with the largest response and the kernel with the smallest response. The maximum-response feature of the kernel with the largest response receives a soft count, meaning that its response value is accumulated, while the minimum-response feature of the kernel with the smallest response receives a hard count, meaning that its count is incremented by one. The feature values of all other kernels remain unchanged at that time point. Accumulating these values across all time points produces the two features associated with each kernel. Overall, Hydra therefore produces $2dgk$ features.

Hydra adapts the set of kernel dilations to the length of the time series $L$, which results in more features for longer series. To control the total number of features, we therefore vary the number of groups $g$. Hydra uses powers-of-two dilations whose effective kernel span fits within the series. Since Hydra uses a fixed kernel length of $9$, the number of dilation values is $ d = \left\lfloor \log_2\left(\frac{L-1}{9-1}\right) \right\rfloor + 1. $ The term +1 is included to account for dilation 1. Given a requested Hydra feature budget $B$, we hence set $ g = \left\lfloor \frac{B}{2kd} \right\rfloor, $ and fix $k=8$ kernels per group, as in the default Hydra configuration.

Hydra operates on both the raw time series and its first-order difference. In our implementation, $\lfloor g/2 \rfloor$ groups are applied to the raw time series and $\lfloor g/2 \rfloor$ groups to the first-order difference. Consequently, when $g$ is odd, the actual number of generated features can be slightly smaller than the requested feature budget. For $g=1$, only the raw time series is considered.

Before classification with the tabular foundation model, we concatenate the feature vectors produced by MultiRocket and Hydra into a single combined representation.

%% file: evaluation.tex
\section{Evaluation}

In this section, we evaluate \method{} on established univariate and multivariate time-series classification benchmarks. We compare our results with those reported in the corresponding benchmark studies, following the same dataset protocols.

\subsection{Experimental Setup}
We adapt the feature budget depending on the number of combined training and test samples, using \(10{,}000\) features for fewer than \(1{,}000\) samples, and \(2{,}000\) in all other cases. We divide the budget equally between Hydra \citep{dempster2023hydra} and MultiRocket \citep{tan2022multirocket} before concatenating their features. We provide further implementation details in Appendix~\ref{apx:implementation}.

\paragraph{Evaluation}
We evaluate predictive performance using accuracy and summarize performance across datasets using mean ranks. In Appendix~\ref{apx:benchmark-results} we assess pairwise differences using two-sided Wilcoxon signed-rank tests with Holm correction \citep{holm1979simple} and report dataset-level wins, draws, and losses.

\subsection{Univariate Evaluation}

We follow the UTF-112 evaluation of \citet{middlehurst2024bake}, comprising 112 univariate datasets and 30 resamples per dataset. We evaluate \method{} on the same resamples\footnote{ The authors provide the resampled datasets at {\url{https://tsml-eval.readthedocs.io/en/stable/publications/2023/tsc_bakeoff/tsc_bakeoff_2023.html} --- Direct link to data: \url{https://drive.google.com/file/d/1V36LSZLAK6FIYRfPx6mmE5euzogcXS83/view?usp=sharing}} } and compare it with the results reported by \citet{middlehurst2024bake}. The comparison includes HIVE-COTE 2.0 (HC2) \citep{middlehurst2021hivecote2}, MR-Hydra \citep{tan2022multirocket,dempster2023hydra}, QUANT \citep{dempster2024quant}, RDST \citep{guillaume2022rdst}, WEASEL-2 \citep{schaefer2023weasel}, FreshPRINCE \citep{middlehurst2022freshprince}, Proximity Forest (PF) \citep{lucas2019proximity}, and H-InceptionTime \citep{fawaz2020inceptiontime}.

In Figure~\ref{fig:utf112-cd-accuracy}, we show the critical-difference plot of the per-dataset accuracy ranks. We observe that \method{} achieves the best (lowest) average rank. However, we find no statistically significant difference between \method{} and HC2. In contrast, \method{} significantly outperforms MR-Hydra. The comparison with MR-Hydra is particularly relevant because it uses the same MultiRocket--Hydra representation family without the TabPFN-3 classifier.

\begin{figure}[t]
	\centering
	\begin{subfigure}[t]{0.49\textwidth}
		\centering
		\includegraphics[width=0.96\linewidth]{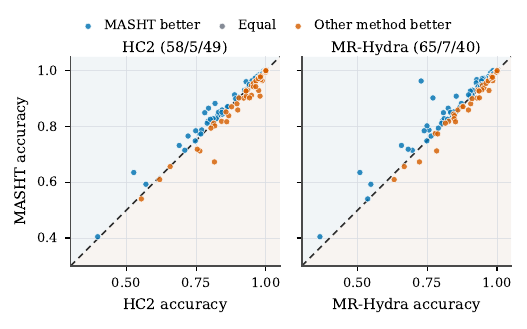}
		\caption{UTF-112.}
		\label{fig:utf112-scatter-comparisons}
	\end{subfigure}\hfill
	\begin{subfigure}[t]{0.49\textwidth}
		\centering
		\includegraphics[width=0.96\linewidth]{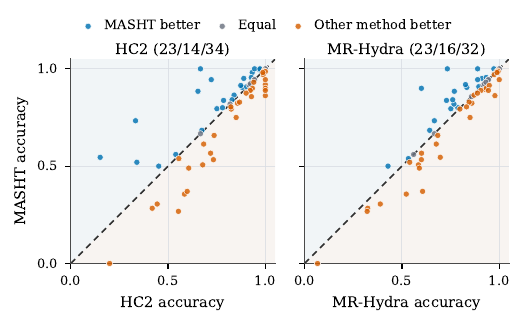}
		\caption{Multiverse.}
		\label{fig:multiverse-scatter-comparisons}
	\end{subfigure}
	\caption{Per-dataset accuracy comparisons of \method{} with HC2 and MR-Hydra
		on (a) UTF-112 and (b) Multiverse. Plot titles report wins/draws/losses for
		\method{}.}
	\label{fig:scatter-comparisons}
\end{figure}
In Figure~\ref{fig:utf112-scatter-comparisons},  we show the per-dataset accuracy of \method{} against HC2 and MR-Hydra and see how these aggregate
differences arise. Against HC2, the dataset-level results cluster closely around
the diagonal, indicating very similar performance without systematically large
differences. The comparison with MR-Hydra is also close for the majority of
datasets, but \method{} performs clearly better on a small number of datasets.
Thus, the improvement over MR-Hydra is concentrated in several pronounced gains
rather than a uniform advantage across the benchmark.

We report the full aggregate metrics and pairwise statistics in Appendix~\ref{apx:univariate-results}, with results in Tables~\ref{tab:utf112-main-results} and~\ref{tab:utf112-pairwise}. These also show that \method{} achieves the highest mean accuracy, balanced accuracy, and AUROC, although its mean accuracy exceeds HC2 by only \(0.001\).

\subsection{Multivariate Evaluation}

For multivariate classification, we follow the Multiverse benchmark of \citet{middlehurst2026multiverse},\!\footnote{We compare methods on all datasets for which results for all methods are available: \url{https://github.com/aeon-toolkit/multiverse/blob/main/results/multiverse/accuracy_mean.csv}} which comprises 71 datasets. We evaluate \method{} on these and compare it with the results that \citet{middlehurst2026multiverse} report. The benchmark covers a broad selection of distance-based, interval, shapelet, dictionary, feature-based, convolutional, hybrid, and deep-learning classifiers, including HC2 \citep{middlehurst2021hivecote2}, MR-Hydra \citep{tan2022multirocket,dempster2023hydra}, QUANT \citep{dempster2024quant}, ROCKET \citep{dempster2020rocket}, Arsenal \citep{middlehurst2021hivecote2}, FreshPRINCE \citep{middlehurst2022freshprince}, CIF \citep{middlehurst2020cif}, DrCIF \citep{middlehurst2020cif}, RDST \citep{guillaume2022rdst}, RIST \citep{middlehurst2023rist}, STC \citep{bostrom2017binary}, TDE \citep{middlehurst2020tde}, Catch22 \citep{lubba2019catch22}, InceptionTime \citep{fawaz2020inceptiontime}, LITETime \citep{ismailfawaz2025litetime}, and 1NN-DTW \citep{shokoohi2017generalizing}.

\begin{figure*}[t]
	\centering
	\includegraphics[width=\textwidth]{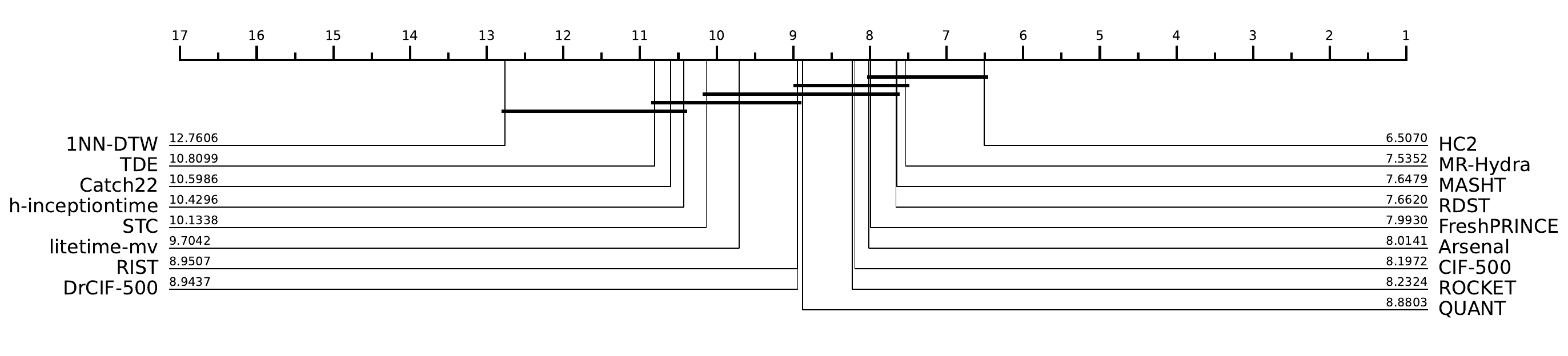}
	\caption{Critical-difference diagram for accuracy on the Multiverse benchmark.}
	\label{fig:multiverse-cd-accuracy}
\end{figure*}

In Figure~\ref{fig:multiverse-cd-accuracy}, we show the corresponding critical-difference diagram. We observe that HC2 obtains the best mean rank and mean accuracy, followed by MR-Hydra, with \method{} close behind. \method{} reaches a mean accuracy of \(0.795\), compared with \(0.805\) for HC2 and \(0.797\) for MR-Hydra. However, the leading methods are not significantly different. Overall, the multivariate results are more modest than the univariate results.

In Figure~\ref{fig:multiverse-scatter-comparisons}, we again examine the per-dataset accuracy against HC2 and MR-Hydra. We observe a wider spread around the diagonal than on UTF-112, indicating larger differences on individual datasets. Against HC2, \method{} wins on 23 datasets, draws on 14, and loses on 34; against MR-Hydra, it wins on 23, draws on 16, and loses on 32.

Based on a manual inspection of the results, we observe particularly strong performance for \method{} on several binary or few-class datasets and short motion datasets with many channels, such as the IRDS and UIPRMD variants. HC2 more often leads on biosignal, gesture, and motion datasets with richer class structure or longer temporal context, including Handwriting, IEEEPPG, EthanolConcentration, and PhotoStimulation.

We report the full aggregate results and pairwise statistics in Appendix~\ref{apx:multivariate-results}, with results in Tables~\ref{tab:multiverse-main-results} and~\ref{tab:multiverse-pairwise}. The mean accuracies of \(0.795\) for \method{}, \(0.805\) for HC2, and \(0.797\) for MR-Hydra support the same conclusion: \method{} is competitive with the strongest methods on Multiverse, but does not establish a new state of the art there.

%% file: resource_experiments.tex
\subsection{Resource Experiments}
\label{sec:resource-experiments}

We complement the benchmark with experiments that vary the resources available to the classifier, the fraction of labeled training cases, the number of transformed features, and the number of TabPFN ensemble members. With ridge regression, we provide a linear baseline that receives exactly the same training cases and MultiRocket–Hydra features as TabPFN. Consequently, the comparison isolates the classifier rather than changes in the temporal representation.

\begin{figure}[t]
	\centering
	\includegraphics[width=\columnwidth]{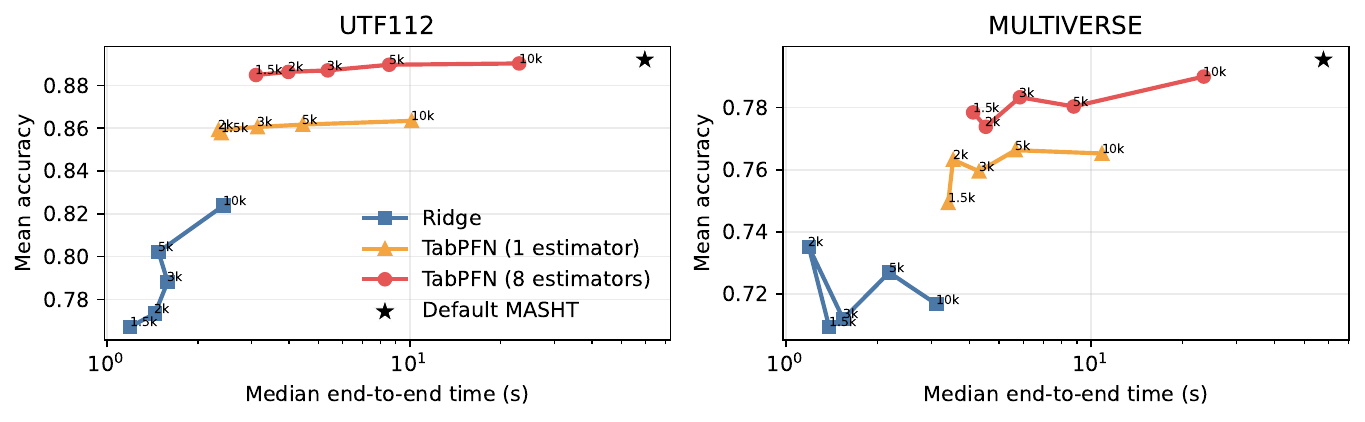}
	\caption{Mean accuracy and median end-to-end time across feature and
		inference budgets. Accuracy uses all datasets; runtime uses the consistently
		measured subset. Annotations give requested feature widths. Stars show the
		published default MASHT aggregates; their original timing protocol differs
		from the controlled resource measurements.}
	\label{fig:resource-pareto}
\end{figure}

\paragraph{Accuracy--Runtime Trade-offs}
To evaluate the accuracy--runtime trade-off, we vary the feature budget $B$ and the number of estimators that \tabpfnIII uses \citep{grinsztajn2026tabpfn3}. In Figure~\ref{fig:resource-pareto}, we show the mean accuracy and the median end-to-end time, that is, the time required for feature transformation, fitting, and prediction. For context, the stars represent the default \method configuration that we presented above, which uses a feature budget of up to $10,000$ and enables automatic estimator scaling in \tabpfnIII (up to 32 estimators). We observe that with 8 estimators the smallest feature budget retains nearly all predictive performance. Increasing $B$ from 1,500 to 10,000 improves mean accuracy by only 0.005 on UTF-112 (0.885 to 0.890) and 0.012 on Multiverse (0.778 to 0.790).

In contrast, median end-to-end time rises from 3.11 to 22.99 seconds and from 4.11 to 23.43 seconds. Thus, wider representations occupy a weak part of the empirical accuracy--runtime frontier. We also observe a clear accuracy difference between the classifiers. At \(B=1{,}500\), TabPFN with 8 estimators achieves 0.118 higher accuracy than the matched Ridge baseline on UTF-112 and 0.069 higher accuracy on Multiverse. Using eight rather than one estimator adds 0.027 and 0.029 accuracy, respectively. TabPFN with one estimator is consequently a useful faster variant, although its accuracy loss is consistent rather than negligible.

\input{Figures/resource/label_fraction}\paragraph{Label Efficiency}
Next, we evaluate how accuracy changes with the number of available training samples. At $B=2{,}000$, we retain class-stratified fractions of the UTF-112 training sets, with at least one example from every class. The minimum-per-class rule makes the realized fraction slightly larger than requested for small datasets. In Figure~\ref{fig:resource-label-efficiency}, we show that the mean accuracy of both classifiers increases monotonically with the label budget, but that TabPFN benefits more sharply between 10\% and 25\%. With a requested label fraction of only 10\% (a median of 23 training cases), TabPFN achieves a mean accuracy of 0.654, compared with 0.589 for Ridge. The largest gains over Ridge occur at 25\% and 50\%, where TabPFN gains 0.118 and 0.120, respectively; the gap narrows only slightly, to 0.113, when all labels are available.

%% file: Figures/resource/label_fraction.tex
\begin{wrapfigure}{R}{0.35\textwidth}
	\centering
	\includegraphics[width=\linewidth]{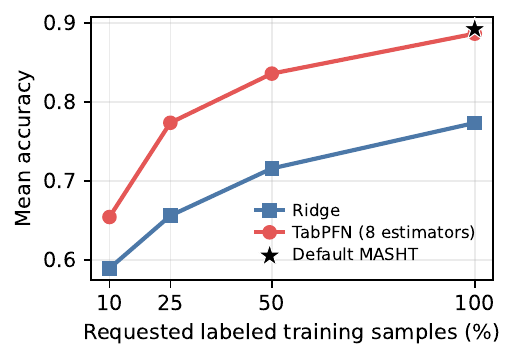}
	\caption{Mean UTF-112 accuracy under varying fractions of labeled training
		samples at $B=2000$.}
	\label{fig:resource-label-efficiency}
\end{wrapfigure}

%% file: conclusion.tex
\section{Discussion and Conclusion}

We presented \method{}, a modular time-series classification approach that combines MultiRocket and Hydra features with TabPFN-3. The contrast between the two benchmarks is the central empirical finding. On UTF-112, \method{} achieves the highest mean accuracy and significantly outperforms MR-Hydra, while showing no statistically significant difference from HIVE-COTE 2.0. On Multiverse, it is competitive with the strongest published methods but does not lead them. This suggests that representing a time series as a compact feature vector is particularly effective in the univariate setting. The comparison with MR-Hydra further indicates that TabPFN-3 is a promising classifier for the MultiRocket--Hydra features. The multivariate results also indicate where the current representation may be less effective. HIVE-COTE 2.0 combines interval, dictionary, shapelet, and convolutional representations, which may capture complementary dependencies between channels and across time.

The resource experiments provide an additional perspective on the role of the classifier and the cost of the representation. When given the same training cases and features, TabPFN outperforms Ridge on both benchmarks. On UTF-112, this advantage also persists across the evaluated label budgets. We further observe that the smallest feature budget with eight estimators retains nearly all predictive performance, while wider representations substantially increase runtime for modest accuracy gains. These results suggest that the feature and inference budgets are practical ways to adjust the accuracy--runtime trade-off.

Overall, these results show that tabular foundation models are promising classifiers for random convolutional time-series features, particularly in the univariate setting. Standardized efficiency comparisons and representations that better capture multivariate temporal structure are the main directions for future work.

%% file: appendix.tex
\appendix

\section{Implementation Details}
\label{apx:implementation}

We implement the benchmark code in Python using \texttt{aeon==1.4.0}, \texttt{tabpfn}, \texttt{scikit-learn}, and PyTorch. The UTF-112 runs process the numbered train/test splits available for each dataset and average metrics across splits. We ran all of our experiments on an NVIDIA A100 GPU with 40~GB of memory. We use TabPFN-3 through the \texttt{tabpfn} package. For the main benchmark runs, we use \texttt{TabPFNClassifier} with \texttt{n\_estimators=8}, and enable automatic estimator scaling. For the controlled resource experiments, we instead use one or eight estimators, and disable automatic estimator scaling. Both configurations use \texttt{fit\_mode=low\_memory}, CUDA execution, automatic inference precision, no tuning configuration, hidden progress bars, and \texttt{ignore\_pretraining\_limits=True}.

The Ridge baseline in the controlled resource experiments uses \texttt{RidgeClassifierCV} from \texttt{scikit-learn}. For each condition, it selects the regularization strength from ten logarithmically spaced values between $10^{-3}$ and $10^{3}$ using the default efficient leave-one-out cross-validation setup (\texttt{cv=None}) on the available labeled training cases. Ridge receives the same training cases and transformed features as the corresponding TabPFN configuration.

\clearpage\section{Detailed Benchmark Results}
\label{apx:benchmark-results}

\subsection{Metrics and Statistical Comparisons}
\label{apx:evaluation-metrics} On UTF-112, we report balanced accuracy and AUROC in addition to accuracy. We compute balanced accuracy as the unweighted mean of the recall for each class, giving every class equal weight regardless of its frequency. For binary datasets, we compute AUROC with the minority class in the training split as the positive class. For multiclass datasets, we compute one-vs-rest AUROC for each class and average these scores using the class frequencies in the training split as weights.

For pairwise accuracy comparisons, we report dataset-level wins, draws, and losses, the mean accuracy difference (\method{} minus the baseline), and both unadjusted and Holm-adjusted \(p\)-values from two-sided Wilcoxon signed-rank tests \citep{holm1979simple}.

\subsection{Univariate Results}
\label{apx:univariate-results} We aggregate UTF-112 results over 30 resamples per dataset. In Table~\ref{tab:utf112-main-results}, we report mean accuracy, balanced accuracy, and AUROC, as well as mean accuracy rank. \method{} achieves a mean accuracy of \(0.892\), which exceeds HC2's \(0.891\) only slightly. In Table~\ref{tab:utf112-pairwise}, we report 58 wins, 5 draws, and 49 losses against HC2, with a mean accuracy difference of \(0.001\); this difference is not significant after Holm correction. Against MR-Hydra, we report 65 wins, 7 draws, and 40 losses, with a mean accuracy difference of \(0.008\), which remains significant after Holm correction.

\subsection{Multivariate Results and Interpretation}
\label{apx:multivariate-results} In Table~\ref{tab:multiverse-main-results}, we report mean accuracies of \(0.795\) for \method{}, \(0.805\) for HC2, and \(0.797\) for MR-Hydra. In Table~\ref{tab:multiverse-pairwise}, we report 23 wins, 14 draws, and 34 losses against HC2, with a mean accuracy difference of \(-0.010\). Against MR-Hydra, we report 23 wins, 16 draws, and 32 losses, with a mean accuracy difference of \(-0.002\). None of the reported pairwise comparisons is significant.

\begin{table}[ht]
	\centering
	\caption{Mean performance on the 112 univariate UTF-112 datasets. We compare
		baseline results from \citet{middlehurst2024bake} with \method{} on the same
		30 resamples. Rank is the mean per-dataset accuracy rank.}
	\label{tab:utf112-main-results}
	\input{generated/evaluation/utf112_main_results.tex}
\end{table}

\begin{table}[ht]
	\centering
	\caption{Pairwise accuracy comparison of \method{} with all compared UTF-112
		baselines. W/D/L denotes the numbers of datasets on which \method{} wins, draws,
		or loses. We compute differences as \method{} minus the baseline.}
	\label{tab:utf112-pairwise}
	\input{generated/evaluation/utf112_pairwise_wdl.tex}
\end{table}

\begin{table}[ht]
	\centering
	\caption{Mean accuracy on the 71 multivariate Multiverse datasets. We use baseline
		results from \citet{middlehurst2026multiverse}. Rank is the mean per-dataset
		accuracy rank.}
	\label{tab:multiverse-main-results}
	\input{generated/evaluation/multiverse_main_results.tex}
\end{table}

\begin{table}[ht]
	\centering
	\caption{Pairwise accuracy comparison of \method{} with selected Multiverse
		baselines. W/D/L denotes the numbers of datasets on which \method{} wins, draws,
		or loses. We compute differences as \method{} minus the baseline.}
	\label{tab:multiverse-pairwise}
	\input{generated/evaluation/multiverse_pairwise_wdl.tex}
\end{table}

\subsection{Runtime}

Runtime provides an important additional perspective on the accuracy results. Our median end-to-end runtime for \method{} on UTF-112 is \(59.7\) seconds per dataset (Table~\ref{tab:runtime}). For comparison, \citet{middlehurst2024bake} report a median HC2 training time of \(15.28\) minutes across 142 univariate datasets, with a total training time of \(263.89\) hours and individual times ranging from \(38.94\) seconds to \(65.66\) hours. Their experiments used an Intel Xeon Gold 5220R CPU, whereas all \method{} experiments were run using an NVIDIA A100 GPU with 40~GB of memory. The reported timing boundaries and dataset collections also differ. The values therefore suggest that \method{} may offer a favorable accuracy--runtime trade-off, but they do not constitute a controlled speed comparison; establishing one requires both methods to be run on the same datasets, hardware, and timing protocol.

\begin{table}[t]
	\centering
	\caption{Runtime summary for default \method{} in seconds per
		dataset.}
	\label{tab:runtime}
	\input{generated/evaluation/runtime_summary.tex}
\end{table}

%% file: generated/evaluation/utf112_main_results.tex
\begin{tabular}{lrrrr}
	\toprule
	Method          & Acc.           & Bal. Acc.      & AUROC          & Rank          \\
	\midrule
	\method{}       & \textbf{0.892} & \textbf{0.872} & \textbf{0.970} & \textbf{3.27} \\
	HC2             & 0.891          & 0.871          & 0.968          & 3.41          \\
	MR-Hydra        & 0.884          & 0.866          & 0.913          & 3.88          \\
	H-InceptionTime & 0.876          & 0.861          & 0.959          & 4.95          \\
	RDST            & 0.876          & 0.856          & 0.907          & 5.11          \\
	WEASEL-2        & 0.874          & 0.853          & 0.905          & 5.37          \\
	QUANT           & 0.867          & 0.845          & 0.962          & 5.42          \\
	FreshPRINCE     & 0.855          & 0.834          & 0.958          & 6.54          \\
	PF              & 0.837          & 0.819          & 0.942          & 7.05          \\
	\bottomrule
\end{tabular}

%% file: generated/evaluation/utf112_pairwise_wdl.tex
\begin{tabular}{lrrrr}
	\toprule
	Baseline        & W/D/L   & Mean $\Delta$ & $p$  & Holm $p$ \\
	\midrule
	HC2             & 58/5/49 & 0.001         & 0.42 & 0.42     \\
	MR-Hydra        & 65/7/40 & 0.008         & 0.01 & 0.03     \\
	H-InceptionTime & 69/4/39 & 0.016         & 0.00 & 0.00     \\
	RDST            & 75/5/32 & 0.016         & 0.00 & 0.00     \\
	WEASEL-2        & 79/3/30 & 0.018         & 0.00 & 0.00     \\
	QUANT           & 86/5/21 & 0.025         & 0.00 & 0.00     \\
	FreshPRINCE     & 97/3/12 & 0.037         & 0.00 & 0.00     \\
	PF              & 94/5/13 & 0.055         & 0.00 & 0.00     \\
	\bottomrule
\end{tabular}

%% file: generated/evaluation/multiverse_main_results.tex
\begin{tabular}{lrr}
	\toprule
	Method          & Acc.           & Rank          \\
	\midrule
	HC2             & \textbf{0.805} & \textbf{6.51} \\
	MR-Hydra        & 0.797          & 7.54          \\
	\method{}       & 0.795          & 7.65          \\
	RDST            & 0.797          & 7.66          \\
	FreshPRINCE     & 0.792          & 7.99          \\
	Arsenal         & 0.792          & 8.01          \\
	CIF-500         & 0.796          & 8.20          \\
	ROCKET          & 0.793          & 8.23          \\
	QUANT           & 0.786          & 8.88          \\
	DrCIF-500       & 0.769          & 8.94          \\
	RIST            & 0.778          & 8.95          \\
	litetime-mv     & 0.750          & 9.70          \\
	STC             & 0.784          & 10.13         \\
	h-inceptiontime & 0.731          & 10.43         \\
	Catch22         & 0.756          & 10.60         \\
	TDE             & 0.759          & 10.81         \\
	1NN-DTW         & 0.683          & 12.76         \\
	\bottomrule
\end{tabular}

%% file: generated/evaluation/multiverse_pairwise_wdl.tex
\begin{tabular}{lrrrr}
	\toprule
	Baseline    & W/D/L    & Mean $\Delta$ & $p$  & Holm $p$ \\
	\midrule
	HC2         & 23/14/34 & -0.010        & 0.15 & 1.00     \\
	MR-Hydra    & 23/16/32 & -0.002        & 0.45 & 1.00     \\
	RDST        & 33/7/31  & -0.002        & 0.84 & 1.00     \\
	QUANT       & 35/14/22 & 0.010         & 0.39 & 1.00     \\
	ROCKET      & 33/9/29  & 0.002         & 0.53 & 1.00     \\
	Arsenal     & 31/11/29 & 0.003         & 0.65 & 1.00     \\
	FreshPRINCE & 35/12/24 & 0.004         & 0.61 & 1.00     \\
	\bottomrule
\end{tabular}

%% file: generated/evaluation/runtime_summary.tex
\begin{tabular}{@{}lrrr@{}}
	\toprule
	Runtime component           & Mean (s)        & Median (s)      & Maximum (s)     \\
	\midrule
	\multicolumn{4}{@{}l}{\textbf{UTF-112}}                                           \\
	\textbf{Total runtime}      & \textbf{43.954} & \textbf{59.743} & \textbf{83.457} \\
	\quad Hydra transform       & 0.939           & 0.456           & 7.884           \\
	\quad MultiROCKET transform & 0.085           & 0.059           & 0.507           \\
	\quad Classification        & 42.928          & 59.528          & 83.124          \\
	\addlinespace[2pt]
	\multicolumn{4}{@{}l}{\textbf{Multiverse}}                                        \\
	\textbf{Total runtime}      & \textbf{47.312} & \textbf{57.813} & \textbf{79.962} \\
	\quad Hydra transform       & 1.177           & 0.375           & 14.173          \\
	\quad MultiROCKET transform & 0.279           & 0.100           & 4.848           \\
	\quad Classification        & 45.854          & 57.144          & 70.312          \\
	\bottomrule
\end{tabular}